%% file: thesis.tex
\documentclass[a4paper, twoside]{report}

\usepackage[T1]{fontenc}
\usepackage[utf8]{inputenc}
\usepackage{lmodern}
\usepackage{microtype}

\usepackage{geometry}
\usepackage{amsmath, amssymb, amsthm, amsfonts, mathtools}
\usepackage{bm}
\usepackage{bbm}
\usepackage{paralist}
\usepackage{etoolbox}
\usepackage[colorinlistoftodos]{todonotes}
\usepackage[most]{tcolorbox}

\usepackage{\string dsfont}

\usepackage[acronym, toc, nonumberlist]{glossaries}
\usepackage{multicol}
{\begin{multicols}{2}\begin{description}}%
{\end{description}\end{multicols}}
\makeglossaries
\input{acronyms}



\usepackage{graphicx}
\usepackage{subcaption}
\usepackage{caption}
\usepackage{wrapfig}
\usepackage{booktabs}
\usepackage{float}
\usepackage{multirow}
\usepackage{tabularx}
\usepackage{adjustbox}
\usepackage{pdflscape}

\usepackage{xcolor}
\definecolor{bordeau}{rgb}{0.3515625,0,0.234375}

\usepackage{algorithm}
\usepackage{algorithmic}

\usepackage{titlesec}
\usepackage{enumitem}
\usepackage{setspace}
\usepackage{array}
\usepackage{lipsum}
\usepackage{pgffor}
\usepackage{afterpage}
\usepackage{textpos}
\usepackage{hyperref}
\usepackage{cleveref}
\usepackage{pdfcomment}

\usepackage{stmaryrd}
\usepackage{pifont, mdframed}

\usepackage{epigraph}
\usepackage{wallpaper}
\usepackage{eso-pic}  % for full-page background images

\input{macros}
\input{theorems}

\usepackage[square, numbers]{natbib}
\setcitestyle{authoryear,round,citesep={;},aysep={,},yysep={;}}
\newcommand{\PhDTitle}{Thesis title. It can extend over several lines (even 4 or 5)} 
\newcommand{\PhDname}{Mouad EL Bouchattaoui} 
\newcommand*{\NumOfChapters}{6}
\newcommand*{\NumOfAppendices}{5}

\hypersetup{
	pdfauthor={\PhDname},
	pdfsubject={Ph.D. thesis manuscript},
	pdftitle={\PhDTitle}
}

\begin{document}

% Turn off page numbers for the very beginning (cover etc.)
\pagenumbering{gobble}

% --- Cover Page Section ---
\begingroup
  \newgeometry{margin=0cm}
  \thispagestyle{empty}
  \input{layout/firstpage.tex}
\restoregeometry
\endgroup

% no extra clear here; next page starts immediately

% --- Restore spacing again, explicitly ---
\onehalfspacing  % Or use \setstretch{1.5} if you want more control
\doublespacing
    \input{layout/resume.tex}

    \color{black}   % <-- reset text color

    \clearpage

    \input{layout/dedication.tex}
    \clearpage
    
    \input{layout/acknowledgements.tex}

    \clearpage

    % Clean break before starting arabic numbering
    % (single-page clear above is enough)
    
    \pagenumbering{arabic}
    \tableofcontents
    \listoffigures
    \listoftables

    \clearpage
    \chapter*{Glossary}
    \addcontentsline{toc}{chapter}{Glossary}
    \begin{small}
    \printglossary[type=\acronymtype,nonumberlist=false]
    \end{small}

    \input{notations_table}

    \foreach \i in {1,2,...,\NumOfChapters}{
        \input{\i}
        \clearpage
    }

    \appendix
    \foreach \i in {3,4,...,\NumOfAppendices}{
        \input{appendices/\i}
        \clearpage
    }

    \bibliography{bibliography.bib}
    % \pagebreak

    % \input{layout/lastpage.tex}

\end{document}

%% file: acronyms.tex
\newacronym{rct}{RCT}{Randomized Controlled Trial}
\newacronym{scm}{SCM}{Structural Causal Model}
\newacronym{po}{PO}{Potential Outcomes}
\newacronym{dag}{DAG}{Directed Acyclic Graph}
\newacronym{ite}{ITE}{Individualized Treatment Effect}
\newacronym{ate}{ATE}{Average Treatment Effect}
\newacronym{dgp}{DGP}{Data-Generating Process}
\newacronym{roi}{ROI}{Return on Investment}
\newacronym{sgdbf}{SGDBF}{Saint-Gobain Distribution Bâtiment France}
\newacronym{crl}{CRL}{Causal Representation Learning}
\newacronym{cdvae}{CDVAE}{Causal Dynamic Variational Autoencoder}
\newacronym{rnn}{RNN}{Recurrent Neural Network}
\newacronym{cpc}{CPC}{Contrastive Predictive Coding}
\newacronym{infomax}{InfoMax}{Information Maximization}
\newacronym{sutva}{SUTVA}{Stable Unit Treatment Value Assumption}
\newacronym{iptw}{IPTW}{Inverse Probability of Treatment Weighting}
\newacronym{cte}{CTE}{Contemporaneous Treatment Effect}
\newacronym{pci}{PCI}{Proximal Causal Inference}
\newacronym{iv}{IV}{Instrumental Variable}
\newacronym{msm}{MSM}{Marginal Structural Model}
\newacronym{tmle}{TMLE}{Targeted Maximum Likelihood Estimation}
\newacronym{glm}{GLM}{Generalized Linear Model}
\newacronym{dml}{DML}{Double Machine Learning}
\newacronym{ipm}{IPM}{Integral Probability Metric}
\newacronym{mmd}{MMD}{Maximum Mean Discrepancy}
\newacronym{dtr}{DTR}{Dynamic Treatment Regime}
\newacronym{ipw}{IPW}{Inverse Probability Weighting}
\newacronym{xai}{XAI}{Explainable Artificial Intelligence}
\newacronym{lstm}{LSTM}{Long Short-Term Memory}
\newacronym{rmsm}{RMSM}{Recurrent Marginal Structural Model}
\newacronym{crn}{CRN}{Counterfactual Recurrent Network}
\newacronym{ct}{CT}{Causal Transformer}
\newacronym{cdc}{CDC}{Counterfactual Domain Confusion}
\newacronym{cgm}{CGM}{Causal Graphical Model}
\newacronym{acate}{ACATE}{Augmented Conditional Average Treatment Effect}
\newacronym{cate}{CATE}{Conditional Average Treatment Effect}
\newacronym{mimic}{MIMIC}{Medical Information Mart for Intensive Care}
\newacronym{cmm}{CMM}{Conditional Markov Model}
\newacronym{dgm}{DGM}{Deep Generative Model}
\newacronym{elbo}{ELBO}{Evidence Lower Bound}
\newacronym{gmm}{GMM}{Gaussian Mixture Model}
\newacronym{kl}{KL}{Kullback–Leibler}
\newacronym{dvae}{DVAE}{Dynamic Variational Autoencoder}
\newacronym{pehe}{PEHE}{Precision in Estimation of Heterogeneous Effects}
\newacronym{mse}{MSE}{Mean Squared Error}
\newacronym{club}{CLUB}{Contrastive Log-ratio Upper Bound}
\newacronym{infonce}{InfoNCE}{Information Noise Contrastive Estimation}
\newacronym{gru}{GRU}{Gated Recurrent Unit}
\newacronym{mi}{MI}{Mutual Information}
\newacronym{mine}{MINE}{Mutual Information Neural Estimator}
\newacronym{pkpd}{PK-PD}{PharmacoKinetic-PharmacoDynamic}
\newacronym{nrmse}{NRMSE}{Normalized Root Mean Squared Error}
\newacronym{hd}{HD}{High-Dimensional}
\newacronym{nmf}{NMF}{Nonnegative Matrix Factorization}
\newacronym{saac}{SAAC}{Spectral Algorithm with Additive Clustering}
\newacronym{symnmf}{SymNMF}{Symmetric Nonnegative Matrix Factorization}
\newacronym{ica}{ICA}{Independent Component Analysis}
\newacronym{h}{H}{Homogeneity}
\newacronym{c}{C}{Completeness}
\newacronym{nmi}{NMI}{Normalized Mutual Information}
\newacronym{ssc}{SSC}{Sparse Subspace Clustering}
\newacronym{nwj}{NWJ}{Nguyen, Wainwright, and Jordan}
\newacronym{gnn}{GNNs}{Graph Neural Networks}

\newacronym{whp}{w.h.p}{with high probability}
\newacronym{sota}{SOTA}{state-of-the-art}
\newacronym{wrt}{w.r.t.}{with respect to}
\newacronym{lhs}{LHS}{left-hand side}
\newacronym{rhs}{RHS}{right-hand side}

%% file: macros.tex
\newcommand\blankpage{%
    \null
    \thispagestyle{empty}%
    \addtocounter{page}{-1}%
    \newpage}
\numberwithin{equation}{chapter}

%% file: theorems.tex
\newcommand{\BlackBox}{\rule{1.5ex}{1.5ex}}

\definecolor{darkred}{rgb}{0.6, 0.0, 0.0}  % Define darkred color
\definecolor{teal}{rgb}{0.0, 0.5, 0.5}  % Teal
\definecolor{grayborder}{gray}{0.6} % dark gray for frame
\definecolor{grayback}{gray}{0.95}  % very light gray for background

\newcounter{assumption}[chapter]
\newcounter{definition}[chapter]
\newcounter{conjecture}[chapter]
\newcounter{proposition}[chapter]
\newcounter{lemma}[chapter]
\newcounter{corollary}[chapter] 
\newcounter{remark}[chapter]
\newcounter{example}[chapter]
\newcounter{theoremBox}[chapter]
\newcounter{question}[chapter]
\renewcommand{\theassumption}{Asm\thechapter.\arabic{assumption}}      % 'Asm' is clearer than 'A'
\renewcommand{\thedefinition}{Def\thechapter.\arabic{definition}}       % 'Def' is concise and common
\renewcommand{\theconjecture}{Conj\thechapter.\arabic{conjecture}}      % 'Conj' is more recognizable than 'CN'
\renewcommand{\theproposition}{Prop\thechapter.\arabic{proposition}}    % 'Prop' is widely used
\renewcommand{\thelemma}{Lem\thechapter.\arabic{lemma}}                 % 'Lem' is shorter and widely accepted
\renewcommand{\thecorollary}{Cor\thechapter.\arabic{corollary}}         % 'Cor' is standard
\renewcommand{\theremark}{Rem\thechapter.\arabic{remark}}               % 'Rem' avoids ambiguity
\renewcommand{\theexample}{Ex\thechapter.\arabic{example}}              % 'Ex' is good as-is
\renewcommand{\thequestion}{Q\thechapter.\arabic{question}}             % 'Q' is fine, but could also use 'Que' for consistency
\renewcommand{\thetheoremBox}{Thm\thechapter.\arabic{theoremBox}}       % 'Thm' is standard
\newtcolorbox{intuitionbox}[1][]{
    colback=gray!15,  % Light gray background
    colframe=white,   % No visible border
    boxrule=0pt,      % Remove frame border
    boxsep=5pt,       % Inner padding
    sharp corners,    % Sharp corners
    before skip=15pt plus 3pt minus 2pt, % Space before
    after skip=15pt plus 3pt minus 2pt,  % Space after
    enhanced,         % Enable better rendering
    overlay unbroken={},
    #1
}

%% file: layout/firstpage.tex
\label{form_first}

%%% Données %%%

\newcommand{\NNT}{2025UPAST116}

\newcommand{\ecodoctitle}{Interfaces : matériaux, systèmes, usages}
\newcommand{\ecodocacro}{INTERFACES}
\newcommand{\ecodocnum}{573}
\newcommand{\PhDspeciality}{Mathématiques appliquées} % <-- vérifier l'intitulé exact dans ADUM
\newcommand{\PhDworkingplace}{CentraleSupélec}
\newcommand{\defenseplace}{Paris-Saclay}             % conforme à la recommandation
\newcommand{\defensedate}{6 novembre 2025}           % format JJ Mois AAAA

\newcommand{\GradSchool}{Sciences de l’ingénierie et des systèmes (SIS)}
\newcommand{\GradSchoolRef}{CentraleSupélec}
\newcommand{\ResearchUnit}{Mathématiques et Informatique pour la Complexité et les Systèmes - EA 4037 (Université Paris-Saclay, CentraleSupélec)} % à vérifier sur le site "signature"

\newcommand{\PhDtitle}{Apprentissage de la causalité pour les données longitudinales}
\newcommand{\PhDtitleTrans}{\textit{Learning Causality for Longitudinal Data}}

%%% Établissement / Institution %%%
% \newcommand{\logoEtt}{blank}
\newcommand{\vpostt}{0.1}
\newcommand{\hpostt}{6}
\newcommand{\logoEt}{CENTSUP}
\newcommand{\vpos}{0.1}
\newcommand{\hpos}{11}

%%% Jury (rôles en français) %%%

\newcommand{\jurynameA}{Gilles Faÿ}
\newcommand{\juryadressA}{Professor, Université Paris-Saclay, CentraleSupélec, MICS}
\newcommand{\juryroleA}{Président du jury}

\newcommand{\jurynameB}{Marianne Clausel}
\newcommand{\juryadressB}{Professeure, Université de Lorraine, groupe SiMul, CRAN}
\newcommand{\juryroleB}{Rapporteure \& examinatrice}

\newcommand{\jurynameC}{Erwan Scornet}
\newcommand{\juryadressC}{Professeur, Sorbonne Université, LPSM et SCAI}
\newcommand{\juryroleC}{Rapporteur \& examinateur}

\newcommand{\jurynameD}{Aurore Lomet}
\newcommand{\juryadressD}{Ingénieure de recherche, CEA, centre de Saclay, laboratoire LIAD}
\newcommand{\juryroleD}{Examinatrice}

\newcommand{\jurynameE}{Paul-Henry Cournède}
\newcommand{\juryadressE}{Directeur de recherche, Université Paris-Saclay, CentraleSupélec}
\newcommand{\juryroleE}{Directeur de thèse}

\newcommand{\jurynameF}{Myriam Tami}
\newcommand{\juryadressF}{Maître de conférences, Université Paris-Saclay, CentraleSupélec, MICS}
\newcommand{\juryroleF}{Co-superviseur de thèse}

\newcommand{\jurynameG}{Benoit Lepetit}
\newcommand{\juryadressG}{Chief Data and Analytics Officer, Saint-Gobain}
\newcommand{\juryroleG}{Co-encadrant industriel}

\newcommand{\jurynameH}{Adil Ahidar}
\newcommand{\juryadressH}{Enseignant-chercheur, École Centrale Casablanca}
\newcommand{\juryroleH}{Membre invité}

\label{layout_first}

\thispagestyle{empty}

\begin{textblock}{5}(0,0)
    \textblockcolour{bordeau}
    \includegraphics[width=5cm]{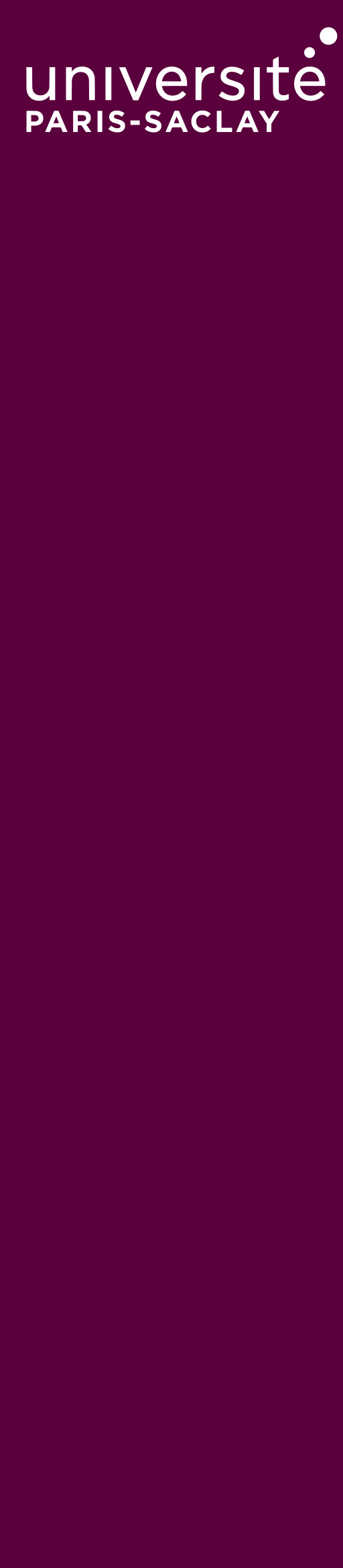}
    \vspace{300mm}
\end{textblock}

\begin{textblock}{3}(0.6,9.5)
    \rotatebox{90}{\textcolor{white}{\fontsize{38}{54}\selectfont Thèse de doctorat}}
\end{textblock}

\begin{textblock}{1}(0.6,3)
	\Large{\rotatebox{90}{\color{white}{NNT : \NNT}}}
\end{textblock}

\begin{textblock}{8}(\hpostt,\vpostt)
	\textblockcolour{white}
	\includegraphics[scale=1]{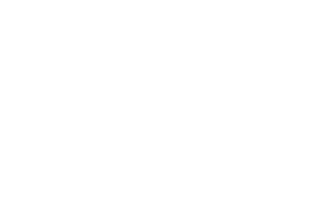}
\end{textblock}

\begin{textblock}{8}(\hpostt,\vpostt)
	\textblockcolour{white}
	\includegraphics[scale=0.4]{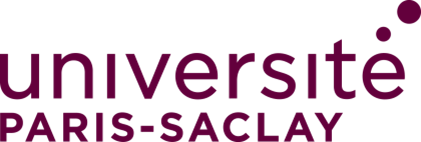}	
\end{textblock}

%% Texte
\begin{singlespace}
\begin{textblock}{10}(5.7,3)
	\textblockcolour{white}
	
	\color{bordeau}
	% ====== TITLE BLOCK (right aligned, unchanged) ======
	\begin{flushright}
		%% Titre principal + traduction
		\huge{\PhDtitle}\\[4pt]
		\Large{\PhDtitleTrans}
		\vfill
		
		\color{black}
		\normalsize
        \vspace{1.5cm}
        \textbf{Thèse de doctorat de l'université Paris-Saclay}
	\end{flushright}

    \vspace{1cm}
    \color{black}
    % ====== SCHOOL / UNIT / DIRECTORS BLOCK (left aligned) ======
    \small
    \begin{flushleft}
        École doctorale n$^{\circ}$\ecodocnum~\ecodoctitle~(\ecodocacro)\\
        Spécialité de doctorat : \PhDspeciality\\
        Graduate School : \GradSchool\\
        Composante référente : \GradSchoolRef\\
        Unité de recherche : \ResearchUnit \\[0.7\baselineskip]
        
        \footnotesize
        Thèse préparée sous la direction de \textbf{\jurynameE} (\juryadressE),\\
        sous la co-supervision de \textbf{\jurynameF} (\juryadressF),\\
        et sous la co-supervision de \textbf{\jurynameG} (\juryadressG).\\[1.2\baselineskip]
		
        Thèse soutenue à \defenseplace, le \defensedate, par\\[0.4\baselineskip]
		
        \Large{\textsc{\PhDname}}
    \end{flushleft}
    \vfill
	
	\color{black}
	%% Jury
	\begin{flushleft}
		\small \textbf{Composition du jury :}\\
	\end{flushleft}
	
	\small
	\newcolumntype{L}[1]{>{\raggedright\let\newline\\\arraybackslash\hspace{0pt}}m{#1}}
	\newcolumntype{R}[1]{>{\raggedleft\let\newline\\\arraybackslash\hspace{0pt}}m{#1}}
	
	\label{jury}
	\begin{flushleft}
	\begin{tabular}{@{} L{9.5cm} R{2.5cm}}
		\jurynameA  \\ \juryadressA & \juryroleA \\[5pt]
		\jurynameB  \\ \juryadressB & \juryroleB \\[5pt]
		\jurynameC  \\ \juryadressC & \juryroleC \\[5pt]
		\jurynameD  \\ \juryadressD & \juryroleD \\[5pt]
		% Ajouter E, F, G, H si la ED l’exige dans la composition affichée
	\end{tabular}
	\end{flushleft}
\end{textblock}
\end{singlespace}

% \afterpage{\blankpage}

%% file: layout/resume.tex
%%%%%%%%%%%%%%%%%%%%%%%%%%%%%%%%%%%%%%%%%%%%%%%%%%%%%%%%%%%%%%%%%%%%%%%%%%%%%%%%%%%%%%%%%%%%%%%%%%%%%%%%%%%%%%%%%%%%%%%%%%%%%%%%%%%%%%%%%%%%%%%%%%%%%%%%%%%%%%%%%%%%%%%
%%% Back cover – Université Paris-Saclay (2 pages: FR then EN)
%%%%%%%%%%%%%%%%%%%%%%%%%%%%%%%%%%%%%%%%%%%%%%%%%%%%%%%%%%%%%%%%%%%%%%%%%%%%%%%%%%%%%%%%%%%%%%%%%%%%%%%%%%%%%%%%%%%%%%%%%%%%%%%%%%%%%%%%%%%%%%%%%%%%%%%%%%%%%%%%%%%%%%%

\label{form_last}

%%% Formulaire / Form %%%

\newcommand{\PhDTitleFR}{Apprentissage de la causalité pour les données longitudinales} % Titre FR
\newcommand{\keywordsFR}{Inférence Causale, Données Longitudinales, Apprentissage de Représentations Causales, Régression Contrefactuelle, Modélisation de Variables Latentes}
% 3 à 6 mots clés, séparés par des virgules

\newcommand{\abstractFR}{%
Cette thèse porte sur l’inférence causale et l’apprentissage de représentations causales (CRL) pour des données de haute dimension évoluant dans le temps, avec des applications en médecine de précision, en marketing et en distribution. La première contribution introduit le modèle Causal Dynamic Variational Autoencoder (CDVAE), conçu pour estimer les effets individuels du traitement (ITE) en capturant l’hétérogénéité non observée dans la réponse au traitement, due à des facteurs de risque latents. Contrairement aux approches classiques supposant que tous les facteurs confondants sont observés, CDVAE se concentre sur les variables latentes influençant la séquence de réponses. Ce modèle repose sur des garanties théoriques relatives à la validité des variables d’ajustement latentes et aux bornes de généralisation de l’erreur d’estimation des ITE. Des évaluations extensives sur des jeux de données synthétiques et réels montrent que CDVAE surpasse les méthodes existantes. De plus, nous démontrons que les modèles à l’état de l’art améliorent significativement leurs estimations d’ITE lorsqu’ils sont enrichis par les substituts latents appris par CDVAE, atteignant des performances proches de l’oracle sans accès direct aux véritables variables d’ajustement.

La deuxième contribution traite le défi de l’estimation des effets de traitement à long terme à travers une nouvelle approche de régression contrefactuelle, axée sur l’efficacité computationnelle et la précision de la prévision. En s’appuyant sur des réseaux de neurones récurrents (RNN) enrichis par le codage prédictif contrastif (CPC) et la maximisation de l’information mutuelle (InfoMax), le modèle capture les dépendances temporelles à long terme en présence de facteurs confondants variant dans le temps, tout en évitant les coûts computationnels associés aux transformers. Ce cadre établit de nouveaux standards en estimation contrefactuelle sur des jeux de données synthétiques et réels, et constitue la première intégration du CPC en inférence causale.

La troisième contribution aborde l’apprentissage de représentations causales (CRL), qui vise à découvrir des facteurs latents causaux influençant des observations de haute dimension. Si les progrès récents du CRL ont mis l’accent sur l’identifiabilité dans l’espace latent, l’interprétabilité de la manière dont ces causes latentes se manifestent dans les variables observées n’est pas encore suffisamment explorée – en particulier dans les données structurées comme les données tabulaires longitudinales. Nous introduisons un mécanisme d’interprétabilité agnostique à l’architecture du modèle utilisé, fondé sur la géométrie du Jacobien du décodeur. En imposant une contrainte de parcimonie auto-expressive sur les colonnes du Jacobien, nous induisons une structure modulaire où les variables observées sont regroupées selon leurs influences latentes partagées. Cette contrainte permet de retrouver des regroupements significatifs – potentiellement chevauchants – reflétant la structure causale sous-jacente. Nous établissons des conditions formelles garantissant l’identification cohérente de ces groupes : dans le cas disjoint, nous prouvons la détection de sous-espaces ; dans le cas chevauchant, nous établissons des bornes d’erreur de classification. Nos résultats montrent que la structure causale latente peut être récupérée à partir de la géométrie différentielle, sans recourir à des hypothèses restrictives telles que les variables ancrées ou les décodages mono-parentaux. Au-delà des apports théoriques, nous proposons une technique de régularisation basée sur le Jacobien, à la fois efficace et adaptée aux domaines à grande dimension.%
}

\newcommand{\PhDTitleEN}{Learning Causality for Longitudinal Data}	% Titre EN
\newcommand{\keywordsEN}{Causal Inference, Longitudinal Data, Causal Representation Learning, Counterfactual Regression, Latent Variable Modeling}
% 3 à 6 mots clés, séparés par des virgules

\newcommand{\abstractEN}{%
This thesis addresses causal inference and causal representation learning (CRL) for high-dimensional, time-varying data, with applications across fields such as precision medicine, marketing, and retail. The first contribution introduces the Causal Dynamic Variational Autoencoder (CDVAE), a model designed to estimate Individual Treatment Effects (ITEs) by capturing unobserved heterogeneity in treatment response due to latent risk factors. Unlike traditional approaches that assume some confounders are unobserved, CDVAE focuses on unobserved variables affecting only the outcome sequence. CDVAE is grounded in theoretical guarantees concerning the validity of latent adjustment variables and generalization bounds on ITE estimation error. Extensive evaluations on synthetic and real-world datasets show that CDVAE outperforms existing baselines. Moreover, we demonstrate that state-of-the-art models significantly improve their ITE estimates when augmented with the latent substitutes learned by CDVAE, approaching oracle-level performance without direct access to the true adjustment variables.

The second contribution expands on the challenge of long-term treatment effect estimation through a novel approach to counterfactual regression over time, prioritizing computational efficiency and long-term forecasting accuracy. By leveraging Recurrent Neural Networks (RNNs) enhanced with Contrastive Predictive Coding (CPC) and Information Maximization (InfoMax), the model captures long-term dependencies in the presence of time-varying confounders while avoiding the computational costs associated with transformers. This framework achieves state-of-the-art results in counterfactual estimation across synthetic and real-world datasets and is the first to incorporate CPC into causal inference.

The third contribution tackles Causal Representation Learning (CRL), which seeks to uncover high-level latent factors that influence complex, high-dimensional observations. While recent advances in CRL have emphasized identifiability in the latent space, they often leave open the interpretability of how latent causes manifest in the observed variables—especially in structured data such as longitudinal tabular records. In this work, we introduce a model-agnostic interpretability layer grounded in the geometry of the decoder’s Jacobian. By enforcing a sparse self-expression prior on the Jacobian columns, we induce a modular structure wherein observed features are grouped according to shared latent influences. This constraint enables the recovery of meaningful, potentially overlapping clusters of observed variables, reflecting the underlying generative graph. We provide formal conditions under which these clusters can be consistently identified: in the disjoint setting, we prove subspace detection guarantees using tools from sparse subspace clustering; in the overlapping case, we derive misclassification bounds under a structured latent-feature model that combines sparsity, incoherence, and dominance conditions. Our results demonstrate that latent-to-observed causal structure can be recovered from gradient geometry without restrictive assumptions such as anchor features or single-parent decoding. Beyond the theoretical contributions, we propose efficient Jacobian-based regularization techniques that scale to high-dimensional domains.%
}

\label{layout_last}

%%%%%%%%%%%%%%%%%%%%%%%%%%%%%%%%%%%%%%%%%%%%%%%%%%%%%%%%%%%%%%%%%%%%%%%%%%%%%%%%%%%%%%%%%%%%%%%%%%%%%%%%%%%%%%%%%%%%%%%%%%%%%%%%%%%%%%%%%%%%%%%%%%%%%%%%%%%%%%%%%%%%%%%
%%% Mise en page / Page layout      
%%%%%%%%%%%%%%%%%%%%%%%%%%%%%%%%%%%%%%%%%%%%%%%%%%%%%%%%%%%%%%%%%%%%%%%%%%%%%%%%%%%%%%%%%%%%%%%%%%%%%%%%%%%%%%%%%%%%%%%%%%%%%%%%%%%%%%%%%%%%%%%%%%%%%%%%%%%%%%%%%%%%%%%

\pagestyle{empty}

%%% ================= PAGE 1 : FRENCH =================
%%% Logo ED
\begin{textblock*}{61mm}(16mm,3mm)
    \textblockcolour{white}
    \noindent\includegraphics[height=24mm]{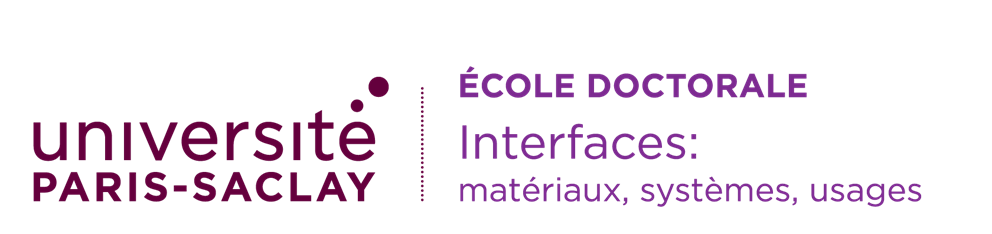}
\end{textblock*}

% ADD THIS to move the white box below the logo:
\vspace*{20mm}  % adjust 20mm up/down until you like the layout

\begin{singlespace}
\begin{center}
\fcolorbox{bordeau}{white}{%
  \parbox{0.95\textwidth}{%
    {\bf Titre :} \PhDTitleFR\par
    \medskip
    {\bf Mots clés :} \keywordsFR\par
    \vspace{-2mm}
    \begin{multicols}{2}
      {\bf Résumé :}\par
      \abstractFR
    \end{multicols}%
  }%
}
\end{center}
\end{singlespace}

\clearpage

%%% ================= PAGE 2 : ENGLISH =================
\pagestyle{empty}

\begin{singlespace}
\begin{center}
\fcolorbox{bordeau}{white}{%
  \parbox{0.95\textwidth}{%
    {\bf Title:} \PhDTitleEN\par
    \medskip
    {\bf Keywords:} \keywordsEN\par
    \vspace{-2mm}
    \begin{multicols}{2}
      {\bf Abstract:}\par
      \abstractEN
    \end{multicols}%
  }%
}
\end{center}

%%% Adresse UPSaclay en bas de la 2e page
\begin{textblock*}{161mm}(10mm,270mm)
  \textblockcolour{white}
  \color{bordeau}
  {\bf\noindent Université Paris-Saclay}\\[1pt]
  \noindent Espace Technologique / Immeuble Discovery\\
  \noindent Route de l’Orme aux Merisiers RD 128 / 91190 Saint-Aubin, France
\end{textblock*}

% \begin{textblock*}{20mm}(182mm,255mm)
%   \textblockcolour{white}
%   \includegraphics[width=20mm]{media/UPSACLAY-petit}
% \end{textblock*}
\end{singlespace}

%% file: layout/dedication.tex
% created on 2019-12-13
% @author : bmazoyer
\newenvironment{dedication}
  {\clearpage           % we want a new page
   \thispagestyle{empty}% no header and footer
   \vspace*{\stretch{1}}% some space at the top 
   \itshape             % the text is in italics
   \raggedleft          % flush to the right margin
  }
  {\par % end the paragraph
   \vspace{\stretch{3}} % space at bottom is three times that at the top
   \clearpage           % finish off the page
  }
  
\begin{dedication}
To my parents. \\
To my lovely sisters. \\
To the kid I was, to the man I am.\\
To the orchid once I watered but withered.
\end{dedication}
\afterpage{\blankpage}

%% file: layout/acknowledgements.tex
\chapter*{Acknowledgements}
\pagenumbering{roman}
\epigraph{Un ouvrage est fini quand on ne peut plus l'améliorer, bien qu'on le sache insuffisant et
incomplet. On en est tellement excédé, qu'on n'a plus le courage d'y ajouter une seule virgule, fût-
elle indispensable. Ce qui décide du degré d'achèvement d'une œuvre, ce n'est nullement une
exigence d'art ou de vérité, c'est la fatigue et, plus encore, le dégoût.}{Emil Cioran}

This thesis would not have been possible without the trust and support of Paul-Henry Cournède and Benoît Lepetit.

I am deeply grateful to Paul-Henry Cournède—my former teacher and my Ph.D. director—for welcoming the idea of a thesis on causality at CentraleSupélec. Your rigorous perspective on my research, your many interventions to ease administrative hurdles, and, above all, your trust in my work have been invaluable. I greatly admire both your intellectual clarity and your human warmth.

Benoît, I vividly remember our first meeting—as if it were yesterday. We had never met before, and our initial contact was simply through LinkedIn. Yet, within the first ten minutes, we were already discussing the framework of a Ph.D. program. Your openness and willingness to engage with a stranger aspiring to pursue a doctorate meant a great deal to me.

I must heartwarmingly acknowledge the support, kindness, and guidance of my Ph.D. supervisor, Myriam Tami. Without you, I would not be the researcher I am today. Working with you has been both a professional privilege and a profoundly human experience. I sincerely cherish your friendship.

This is also a moment to honor all the extraordinary teachers I have had—from primary school onward. I cannot name you all here, but I am profoundly indebted to you for the skills, knowledge, and way of seeing the world that you have instilled in me.

Finally, I must thank the people without whom I would not have seen the light of day or written this thesis: my parents. It is impossible to enumerate the debts I owe you. The sacrifices you have made are engraved in my soul. This thesis is for you.

%% file: notations_table.tex
% ...existing code...
\begin{table}[ht]
\centering
\caption{Summary of Main Mathematical Notations}
\begin{tabular}{ll}
\toprule
\textbf{Notation} & \textbf{Description} \\
\midrule
% Sequence and indices
$T$                   & Length of the longitudinal sequence (number of time steps) \\
$N$                   & Number of individuals \\
$t$                   & Time index, $t = 1, \ldots, T$ \\
$i$                   & Individual (subject) index \\[0.5ex]
$\rho$                & Forecasting horizon or number of lags in an autoregressive model \\

% Observed covariates, treatments, outcomes
$\mathbf{X}_t$        & Time‐varying covariates at time $t$ \\
$\mathbf{V}$          & Static covariates \\
$W_t$                 & Treatment assignment at time $t$ (binary) \\
$Y_t$                 & Observed outcome at time $t$ \\[0.5ex]

% Histories and sequences
$\mathbf{H}_t$        & History/context up to $t$: $(\mathbf{X}_{\le t},\,W_{<t},\,Y_{<t})$ \\
$\mathbf{F}_{t+j}$    & Future components at time $t+j$: $[\mathbf{V},\mathbf{X}_{t+j},W_{t+j-1},Y_{t+j-1}]$ \\
$\mathbf{C}_t^{\mathrm{enc}}$ & Encoded context representation (history) \\[0.5ex]

% Latent and adjustment variables
$\mathbf{U}$          & Unobserved static adjustment variables (risk factors) \\
$\mathbf{Z}$          & Generic latent variable \\[0.5ex]

% Causal‐effect notation
$\tau_t(\cdot)$       & Conditional average treatment effect (CATE) at time $t$ \\
$\tau_t(\mathbf{h}_t,\mathbf{u})$ & Augmented CATE (ACATE) at time $t$ \\[0.5ex]

% Models and parameters
$\Phi$                & Representation function / encoder \\
$f$                   & Outcome model / decoder \\
$\theta,\,\phi$       & Model parameters: generative ($\theta$), inference ($\phi$) \\
$\sigma$              & Variance (noise) parameter in generative model \\[0.5ex]

% Probabilistic notation
$p(\cdot)$            & Probability distribution or density \\
$q_\phi(\cdot)$       & Variational (approximate) posterior distribution \\
$\mathcal{D}_T$       & Observed data up to time $T$ \\[0.5ex]

% Training, objectives & losses
$\mathrm{ELBO}$       & Evidence Lower Bound (variational objective) \\
$\mathcal{L}^{(\mathrm{InfoNCE})}$ & InfoNCE contrastive loss \\
$\mathcal{L}^{(\mathrm{InfoMax})}$ & InfoMax contrastive loss \\
$\mathcal{L}^{\mathrm{CPC}}$ & CPC loss (average InfoNCE) \\
$\mathcal{L}_Y$       & Outcome prediction loss \\
$\mathcal{L}_W$       & Treatment prediction loss \\
$\mathcal{B}$         & Batch of samples \\[0.5ex]

% Metrics & operators
$\mathrm{IPM}$        & Integral Probability Metric (e.g., Wasserstein) \\
$I(\cdot,\cdot)$      & Mutual information \\
$I_{\mathrm{CLUB}}(\cdot,\cdot)$ & CLUB upper bound on MI \\
$\alpha(\cdot)$       & Importance sampling / weighting function \\
$\epsilon$            & Generic error / noise parameter \\[0.5ex]

% Expectations & distributions
$\mathbb{E}$          & Expectation operator \\
$\mathbb{P}$          & Probability operator \\
$\mathcal{N}$         & Normal (Gaussian) distribution \\[0.5ex]

% CRL / subspace clustering
$\mathcal{S}_m$       & $m$-th subspace in union-of-subspaces model \\
$n_m$                 & Number of observed features in subspace $\mathcal{S}_m$ \\
$d_x$                 & Number of observed variables \\
$d_z$                 & Number of latent variables \\
$d_v$                 & Dimensionality of static covariates \\
$\operatorname{aff}(\mathcal{S}_m,\mathcal{S}_{m'})$ & Affinity between subspaces \\
$C \in \mathbb{R}^{d_x\times d_x}$ & Self‐expression coefficient matrix \\
$Df(\cdot)$           & Jacobian of $f$ at $\mathbf{z}$ (shape $d_x\times d_z$) \\
$\nabla_{\mathbf{z}} f(\cdot)$ & Gradient of $f_j$ w.r.t.\ $\mathbf{z}$ (Jacobian column) \\
$Pa(i)$               & Parent set of observed variable $i$ \\
$Ch(j)$               & Children set of latent variable $j$ \\
\bottomrule
\end{tabular}
\label{tab:notations}
\end{table}